\documentclass[fleqn,10pt]{wlscirep}
\usepackage[utf8]{inputenc}
\usepackage[T1]{fontenc}
\usepackage{lineno}
\usepackage{booktabs}
\usepackage{multirow}
\usepackage{adjustbox}
\usepackage[numbers]{natbib}
\usepackage{wrapfig}
\usepackage{enumitem}
\usepackage{multirow}
\usepackage{graphicx}
\usepackage{subcaption}
\usepackage{colortbl}
\usepackage{mathtools}
\usepackage{tikz}
\usepackage{float}
\usepackage{placeins,booktabs,multirow,caption}

\definecolor{blue-violet}{rgb}{0.54, 0.17, 0.89}
\definecolor{bluegray}{rgb}{0.4, 0.6, 0.8}
\definecolor{bleudefrance}{rgb}{0.19, 0.55, 0.91}
\definecolor{darkblue}{rgb}{0.0, 0.0, 0.55}
\definecolor{denim}{rgb}{0.08, 0.38, 0.74}
\definecolor{mediumblue}{rgb}{0.0, 0.0, 0.8}
\definecolor{naplesyellow}{rgb}{0.98, 0.85, 0.37}
\definecolor{carminered}{rgb}{1.0, 0.0, 0.22}
\definecolor{crimsonglory}{rgb}{0.75, 0.0, 0.2}
\definecolor{lightgreen}{rgb}{0.56, 0.93, 0.56}
\definecolor{lightapricot}{rgb}{0.99, 0.84, 0.69}
\definecolor{pastelred}{rgb}{1.0, 0.41, 0.38}
\definecolor{periwinkle}{rgb}{0.8, 0.8, 1.0}
\definecolor{xgreen}{rgb}{0.88, 0.95, 0.83}
\definecolor{uared}{rgb}{0.85, 0.0, 0.3}
\definecolor{debianred}{rgb}{0.84, 0.04, 0.33}
\definecolor{red(ncs)}{rgb}{0.77, 0.01, 0.2}
\definecolor{blush}{rgb}{0.87, 0.36, 0.51}
\definecolor{non-photoblue}{rgb}{0.64, 0.87, 0.93}
\definecolor{lightcornflowerblue}{rgb}{0.6, 0.81, 0.93}
\definecolor{cadmiumgreen}{rgb}{0.0, 0.42, 0.24}
\definecolor{dollarbill}{rgb}{0.52, 0.73, 0.4}
\definecolor{darkgreen}{rgb}{0.0, 0.2, 0.13}
\newcommand\chqsumm{\texttt{CHQ-Summ}}

\newcommand\meqsum{\textsc{MeQSum}}
\title{A Dataset and Benchmark for Consumer Healthcare Question Summarization}
\author[1*]{Abhishek Basu}
\author[2]{Deepak Gupta}
\author[2]{Dina Demner-Fushman}
\author[1]{Shweta Yadav}
\affil[1]{Department of Computer Science, University of Illinois at Chicago, Chicago, IL, USA}
\affil[2]{U.S. National Library of Medicine, National Institutes of
Health, Bethesda, MD, USA}

\begin{abstract}
The quest for seeking health information has swamped the web with consumers' health-related questions. Generally, consumers use overly descriptive and peripheral information to express their medical condition or other healthcare needs, contributing to the challenges of natural language understanding. One way to address this challenge is to summarize the questions and distill the key information of the original question. 
Recently, large-scale datasets have significantly propelled the development of several summarization tasks, such as multi-document summarization and dialogue summarization. However, a lack of domain-expert annotated dataset for the consumer healthcare questions summarization task inhibits the development of an efficient summarization system.
To address this issue, we introduce a new dataset \chqsumm{}{} that contains $1507$ domain-expert annotated consumer health questions and corresponding summaries. The dataset is derived from the community question answering forum and therefore provides a valuable resource for understanding consumer health-related posts on social media.
We benchmark the dataset on multiple state-of-the-art summarization models to show the effectiveness of the dataset.

\end{abstract}
\begin{document}

\raggedbottom
\maketitle



\section*{Background \& Summary}
Healthcare consumers often query the web to find a quick and reliable answer to their healthcare information needs. On average, six million people in the United States alone seek health-related information on the Internet every day \cite{foxwebsite}. One way to facilitate such information-seeking activities is to build a natural language question-answering (QA) system that extracts precise answers from a wide range of health-related information sources. Though existing search engines respond to the general health-related queries to some extent, users often reach out to specialized medical websites or online health communities to seek personalized, high-quality, and trustworthy answers to their complex health questions.
The overly descriptive nature of the questions that contain too much peripheral information brings an additional challenge to the task of automatic analysis and understanding of the questions. Often, such peripheral details are not required to obtain relevant answers, and their removal can lead to significant improvement in QA performance. Therefore, there is a need to develop  {\em automatic question simplification/summarization techniques} before retrieving answers to the questions.

Automatic text summarization is a non-trivial task in natural language processing that aims to generate human-readable, concise text containing salient information of the original document.
The recent development in large-scale neural language models \cite{devlin-etal-2019-bert,raffel2020exploring} has led to
significant performance improvement on several summary generation tasks ( called abstractive summarization), partially due to the availability of large-scale human-annotated training data. The majority of the current summarization datasets are either based on the news articles (e.g., CNN/Dailymail\cite{nallapati-etal-2016-abstractive} and Multi-News\cite{fabbri2019multi} datasets, where headlines are treated as summaries) or the scientific literature ( e.g., PubMed\cite{cohan-etal-2018-discourse}, BioASQ\cite{tsatsaronis2015overview} datasets, where abstracts of the articles serve as summaries). 

\begin{figure}[!t]
  \centering
  \includegraphics[width=0.6\linewidth]{Fig5-1.png}
  \caption{Sample consumer healthcare question and reference summary with the annotated question focus and question type.}
  \label{fig1}
\end{figure}

While significant efforts have been made in open-domain summarization, there are only a few works \cite{yadav2022question,abacha2019summarization,yadav2021nlm,yadav2022towards,yadav2023chqsumm,yadav2021reinforcement,agarwal2025overview,chaturvedi2024aspect} that summarize CHQs. This is partly due to the limited availability of human-annotated training datasets. 
Recently, a manually annotated dataset, \meqsum{} \cite{abacha2019summarization} was created that consists of $1000$ CHQs and their corresponding summaries. A related dataset, MEDIQA-AnS  \cite{savery2020question} focuses on answer summarization for consumer health question answering. While these datasets enabled the research in question answering, the amount of training data remain relatively small, hindering the progress in developing an accurate and efficient CHQ summarization system.

Towards this, we introduce a new CHQ summarization dataset: \chqsumm{}{} that consists of $1507$ domain-expert annotated question-summary pairs from the Yahoo community question answering forum. We advanced the existing dataset in two ways: \textbf{(i)} the dataset is created from the community question answering forum, having a more diverse set of users' questions, and \textbf{(ii)} our \chqsumm{}{} dataset contains additional annotations about question focus and question type of the original question. The question focus is the primary entity of the question and the question type is the aspect of interest about the focus (\textit{c.f.} Figure \ref{fig1}).
Beyond improving question understanding, \chqsumm{} facilitates the development and benchmarking of end-to-end healthcare question answering systems, retrieval-augmented generation (RAG) pipelines, and large language model fine-tuning for consumer health applications \citep{abacha2017overview, yadav2021nlm}. The dataset captures a diverse linguistic spectrum from grammatically incorrect, informal, multi-sentence consumer queries to concise medical terminology, reflecting the variability found in real-world health forums \citep{benabacha2019medqa,abacha2019summarization}. Unlike \meqsum{} \citep{abacha2019summarization}, which focuses primarily on expert-curated summaries, \chqsumm{} leverages community-sourced questions that exhibit greater lexical variety and ambiguity, thereby serving as a more challenging and representative benchmark for question summarization in the healthcare domain. We envision \chqsumm{} as a foundational resource for advancing consumer health NLP, supporting both supervised learning and zero- or few-shot evaluation of emerging large language models in healthcare question answering \citep{devlin-etal-2019-bert,raffel2020exploring}.

\section*{Methods}
\subsection*{Dataset Creation} We utilized the Yahoo! Answers L6 corpus \cite{yahoo_l6} that provides community question answering threads containing users' questions on multiple diverse topics and the answers submitted by other users.
The corpus consists of $4.4$ million question-answer threads with additional metadata information such as the question category, question sub-category, question language, and best answer. Each of the questions has a question title and question content. The question title is the subject line of the original question, and the question content describes the original question with the full description. Since our goal is focused on the consumer healthcare domain, we selected the questions from the ``\textit{Healthcare}" question category. However, to ensure that there are \textbf{(i)} no false positives, and \textbf{(ii)} the questions are from diverse health categories, we devised multiple heuristic-based filtering strategies to filter out irrelevant questions, discussed as follows: 
\begin{figure} 
\centering
\includegraphics[width=0.7\textwidth]{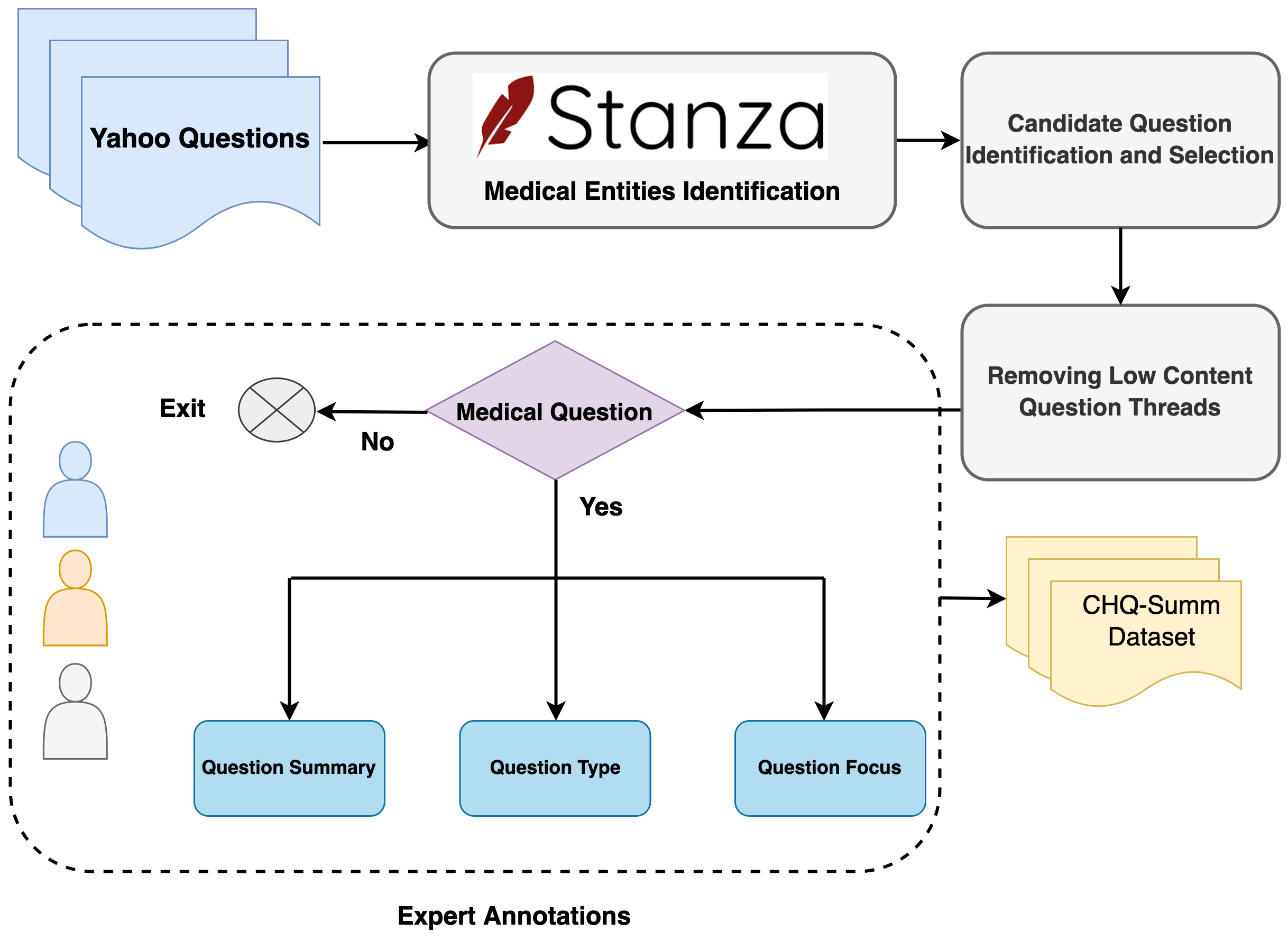}
\caption{The schematic workflow of the \chqsumm{} dataset creation.}
\label{approach}
\end{figure}

\begin{itemize}[noitemsep]
    \item \textbf{Medical Entities Identification:} The very first stage of filtration is to recognize the medical entities present in the question content and title. 
    We utilized Stanza \cite{zhang2021biomedical} biomedical and clinical model trained on the NCBI-Disease corpus 
    for identifying medical entities in the question title and content.

    %
    \item \textbf{Candidate Question Identification and Selection:} 
    In the second stage, we select the question threads having at least one medical entity present in either the question title or the question content. We collected $22,257$ question threads from Yahoo! Answers corpus through this process.
    

\item \textbf{Removing Low Content Question Threads:} In the final stage of filtration, to preserve the quality of our dataset, we further filter out the questions that are short and do not require summarization. 
 Specifically, we removed the question threads with a question content having less than $10$ words. Later, we concatenate the question title and content to form the original question, then ask annotators to formulate the question summary. We call this manually generated question-summary pair \chqsumm{}{}. These pairs constitute the consumer health question summarization dataset. The schematic workflow for creating the \chqsumm{} dataset is depicted in Figure \ref{approach}.

\end{itemize}

\subsection*{Expert Annotations} A team of six annotators ($4$ experts in medical informatics and $2$ experts in medical informatics and medicine) independently annotated  parts of the \chqsumm{}{} dataset. Each annotator was provided with the annotation guidelines and the annotation interface to summarize the CHQs. Given the original question, the annotators were instructed to:
\begin{enumerate}
    \item \textbf{Identify the valid medical question:} A question can be categorized as a valid medical question if it belongs to the following categories: 
    \begin{enumerate}[noitemsep]
        \item Diseases \& conditions, including symptoms
        \item Drugs \& treatments
        \item Medical tests or medical diagnostic \& therapeutic procedures
        \item Other relevant medical topics 
    \end{enumerate}
    \item \textbf{Formulate an abstractive summary:} The abstractive summary of the given question is formulated by keeping the minimum and key information required to answer the question. 
    \item \textbf{Identify the question focus:} 
    Given the effectiveness of the question focus for the consumer health question summarization task, we ask the annotators to annotate each question with the focus term. The question focus is the named entity that is the central theme of the question.
   Generally, a question has a single focus, but for some questions which are about the relation between two drugs or diseases, multiple question foci are allowed. For example, consider Fig. \ref{fig1} where the question focus is \textit{`cataract'}. 
    
    \item \textbf{Identify the question type:} Question type represents some aspect of the question focus that triggered the question and needs to be answered, e.g., in Fig. \ref{fig1} the expecting mother is asking if her child will be susceptible to cataracts more than an average baby. 
    Existing studies \cite{yadav2022question} show that question-type information can guide the model to generate more factually correct summaries, therefore, we annotated each question with the appropriate question types.
    Specifically, we instructed the annotators to classify each question into one of the $33$ question types identified in the previous study \cite{kilicoglu2018semantic}.
    The annotators were expected to select all the question types that could be inferred from the given question. For example, in the Fig. \ref{fig1}, the question type is \textit{`Susceptibility'}.




\end{enumerate}

\subsection*{Annotators and Inter-annotation Agreement} 
We began the annotation process by developing the annotation guidelines and interface. In the first round, a set of $3$ annotators with expertise in medical informatics and medicine annotated the same set of $150$ questions to compute the inter-annotation agreement. In the second round, $3$ more annotators with a medical informatics background were included in the annotation task. For the final annotation, a total of $6$ annotators independently annotated the collection of $500$ questions, of which $30$ were common across all annotators. We computed the inter-annotator agreement (IAA) on the $30$ common question. 

For computing the IAA for the summarization task, we used the ROUGE-L score. In the case of question-focus and question-type identification tasks, we used the F1-score to measure the IAA. We have reported the average ROUGE-L and F1-score amongst all the annotator pairs in Table-\ref{tab:IAA}.
The average IAA for question focus and question types was $83.22$ and $85.44$ (substantial agreement), respectively, on $30$ common questions. On the summarization task, we observed that even though the summary was semantically correct, the average IAA was only $52.8$ (moderate agreement). This was because the annotators' summaries were diverse and did not follow the same word order, even though they were all semantically correct. Since ROUGE-L does not take into account the semantic similarity between two summaries, the ROUGE-L score was not high among annotators. For example, consider Figure-\ref{fig1}, where both variations of the question ask the same thing but do not have too many words in common.


\begin{table}[]
\centering
\resizebox{0.7\textwidth}{!}{%
\begin{tabular}{c|c|c|c}
\hline
\textbf{Samples/Annotators} &
  \textbf{\begin{tabular}[c]{@{}c@{}}Question-focus\\  (F1-score)\end{tabular}} &
  \textbf{\begin{tabular}[c]{@{}c@{}}Question-type\\ (F1-score)\end{tabular}} &
  \textbf{\begin{tabular}[c]{@{}c@{}}Question-summary\\  (ROUGE-L)\end{tabular}} \\ \hline
150/3 &
  85.84 &
  86.28 &
  53.23 \\ 
30/6 &
  83.22 &
  85.44 &
  52.8 \\ \hline
\end{tabular}%
}
\caption{Average inter-annotator agreement score among all the annotators on question-focus recognition, question-type identification, and question-summary generation task.}
\label{tab:IAA}
\end{table}

\begin{figure}[h]
\centering
\begin{subfigure}{.5\textwidth}
\centering
\includegraphics[scale=0.6]{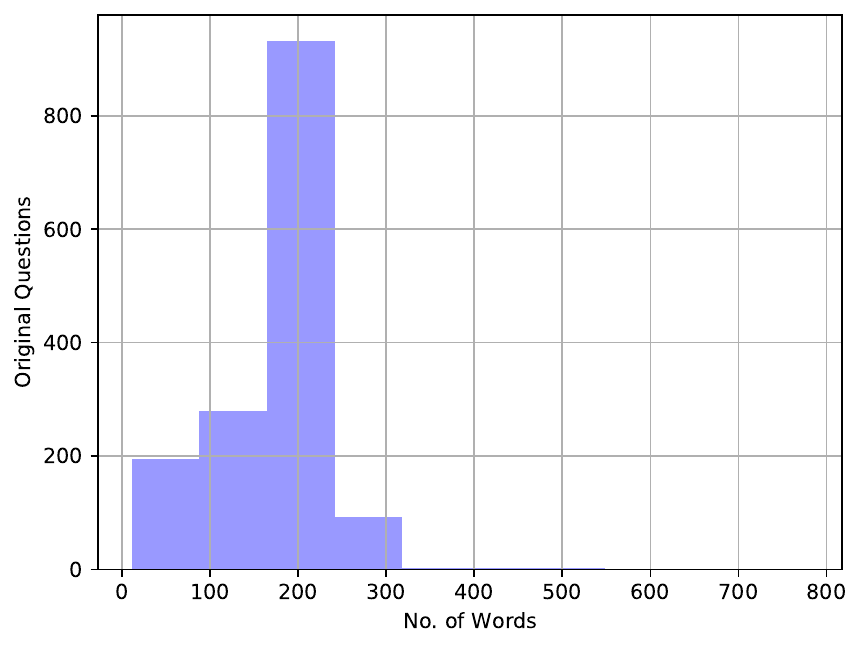}
\caption{}
\end{subfigure}%
\begin{subfigure}{.5\textwidth}
\centering
    \includegraphics[scale=0.6]{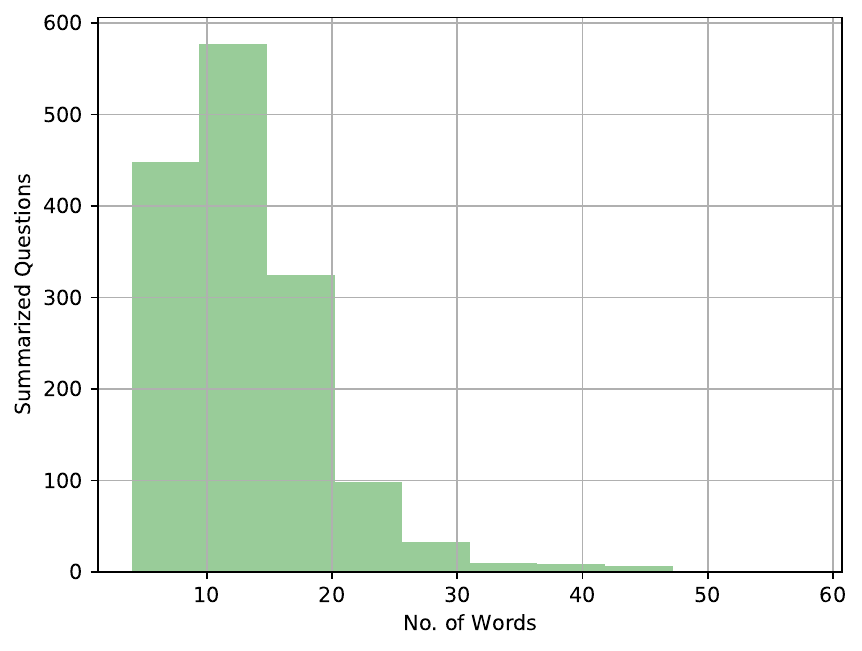}
    \caption{}
\end{subfigure}%
    \caption{Distribution of words in original questions \textbf{(a)} and summarized questions \textbf{(b)} in \chqsumm{} dataset.}
\label{fig:word_distribution}
\end{figure}



\begin{figure*}[t]
\begin{minipage}{.6\textwidth}
\centering
\resizebox{\textwidth}{!}{%
\begin{tabular}{lc|c|c|c|c}
\hline \hline
\multicolumn{2}{c|}{\textbf{Statistics}} &
  \textbf{\begin{tabular}[c]{@{}c@{}}Train\\ (Sentence/Word)\end{tabular}} &
  \textbf{\begin{tabular}[c]{@{}c@{}}Test\\ (Sentence/Word)\end{tabular}} &
  \textbf{\begin{tabular}[c]{@{}c@{}}Validation\\ (Sentence/Word)\end{tabular}} &
  \textbf{\begin{tabular}[c]{@{}c@{}}Total\\ (Sentence/Word)\end{tabular}} \\ \hline
\multicolumn{1}{l|}{\multirow{5}{*}{\textbf{Question}}} & \textbf{Samples}            & 1000         & 400         & 107         & 1507         \\ 
\multicolumn{1}{l|}{}                                   & \textbf{Minimum}            & 1/11         & 2/14        & 2/12        & 1/11         \\ 
\multicolumn{1}{l|}{}                                   & \textbf{Maximum}            & 32/779       & 35/628      & 23/260      & 35/779       \\ 
\multicolumn{1}{l|}{}                                   & \textbf{Mean}               & 10.04/177.21 & 9.99/177.79 & 9.94/168.58 & 10.02/176.75 \\  
\multicolumn{1}{l|}{}                                   & \textbf{Standard Deviation} & 4.54/63.20   & 4.69/68.64  & 4.59/64.01  & 4.58/64.74   \\ \hline
\multicolumn{1}{l|}{\multirow{5}{*}{\textbf{Summary}}}  & \textbf{Samples}            & 1000         & 400         & 107         & 1507         \\
\multicolumn{1}{l|}{}                                   & \textbf{Minimum}            & 1/4          & 1/4         & 1/4         & 1/4          \\
\multicolumn{1}{l|}{}                                   & \textbf{Maximum}            & 2/58         & 2/47        & 2/45        & 2/58         \\ 
\multicolumn{1}{l|}{}                                   & \textbf{Mean}               & 1.02/13      & 1.03/13.62  & 1.02/13.88  & 1.02/13.23   \\  
\multicolumn{1}{l|}{}                                   & \textbf{Standard Deviation} & 0.15/6.03    & 0.18/6.09   & 0.16/7.16   & 0.16/6.14    \\ \hline
  \hline
\end{tabular}%
}
\captionof{table}{Dataset statistics for consumer health question and its corresponding summary in train, test and validation set.}
\label{tab:question-summary-stats}
\end{minipage}
\centering
\quad
\begin{minipage}{.35\textwidth}
\centering
\resizebox{\textwidth}{!}{%
\begin{tabular}{lc|c|c|c|c}
\hline \hline
\multicolumn{2}{c|}{\textbf{Statistics}} &
  \textbf{\begin{tabular}[c]{@{}c@{}}Train\end{tabular}} &
  \textbf{\begin{tabular}[c]{@{}c@{}}Test\end{tabular}} &
  \textbf{\begin{tabular}[c]{@{}c@{}}Validation\end{tabular}} &
  \textbf{\begin{tabular}[c]{@{}c@{}}Total\end{tabular}} \\ \hline
\multicolumn{1}{l|}{\multirow{4}{*}{\textbf{Question Type}}}  & \textbf{Minimum} & 1              & 1             & 1                   & 1              \\ 
\multicolumn{1}{l|}{}                  & \textbf{Maximum}            & 4    & 3    & 3    & 4    \\ 
\multicolumn{1}{l|}{}                  & \textbf{Mean}               & 1.17 & 1.19 & 1.14 & 1.18 \\ 
\multicolumn{1}{l|}{}                  & \textbf{Standard Deviation} & 0.48 & 0.49 & 0.44 & 0.48 \\ \hline
\multicolumn{1}{l|}{\multirow{4}{*}{\textbf{Question Focus}}} & \textbf{Minimum} & 1              & 1             & 1                   & 1              \\ 
\multicolumn{1}{l|}{}                  & \textbf{Maximum}            & 20   & 21   & 9    & 21   \\ 
\multicolumn{1}{l|}{}                  & \textbf{Mean}               & 1.93 & 2.02 & 1.82 & 1.95 \\ 
\multicolumn{1}{l|}{}                  & \textbf{Standard Deviation} & 1.63 & 1.79 & 1.33 & 1.65 \\ \hline  \hline
\end{tabular}%
}
\captionof{table}{Dataset Statistics about question-type and question-focus in train, test and validation set. }
\label{tab:focus-type-stats}
\end{minipage}
\end{figure*}

\section*{Data Records}
We have archived three data records on the Open Science Framework (OSF) \cite{osf_chqsumm}. The OSF link contains a directory \chqsumm{} which stores the developed \chqsumm{} dataset. The \chqsumm{} directory contains the training, validation, and test split of the \chqsumm{} dataset. We also release a README file that contains detailed information about each split, including sample code to process and load the data. The Yahoo! Answers L6 dataset was obtained under a research-only agreement through the Yahoo! Research Alliance Webscope \cite{yahoo_webscope}. Due to Yahoo's copyright issues, we cannot directly share the original Yahoo questions. However, these questions are publicly available through the Yahoo! Answers L6 dataset and can be accessed after signing the required Yahoo agreements \cite{yahoo_l6}. We have provided a Python script to process the Yahoo! Answers L6 dataset and extract the original question corresponding to each sample of the \chqsumm{} dataset \cite{chqsumm_script}. The script also generates the machine-readable inputs to reproduce the benchmark experiments. We use JSON to store the dataset split. Each item in the JSON file is a dictionary with relevant key-value pairs. The keys are \textit{id}, \textit{human\_summary}, \textit{question\_focus}, \textit{question\_type} and \textit{MeSH\_mapping}. The detailed statistics of the \chqsumm{} dataset are shown in Table \ref{tab:question-summary-stats} and Table \ref{tab:focus-type-stats}. The details of each key are provided in the README file. 

\begin{figure}[]
\centering
\begin{subfigure}{.48\textwidth}
\centering
  \includegraphics[scale=0.43]{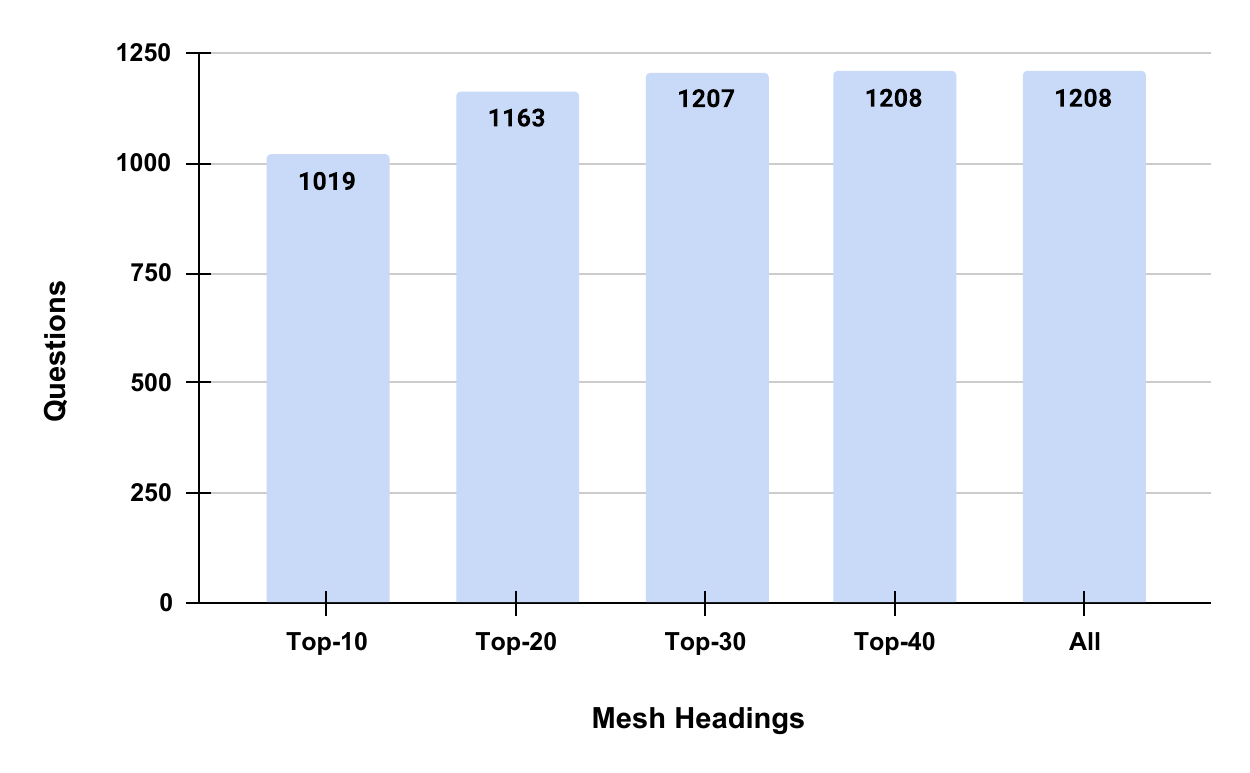}
\caption{}
\end{subfigure}%
\begin{subfigure}{.48\textwidth}
\centering
  \includegraphics[scale=0.43]{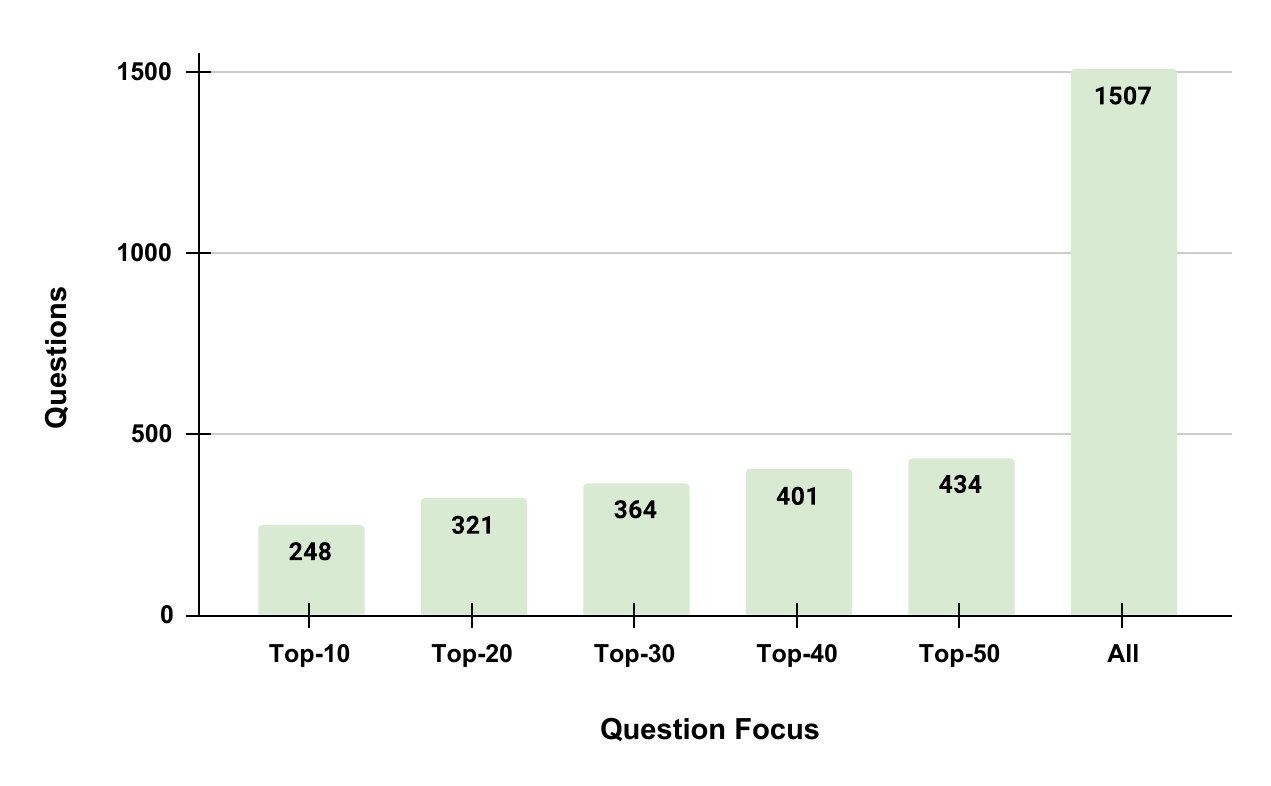}
    \caption{}
\end{subfigure}%
\caption{Coverage of the MeSH headings (\textit{left}) and question focus terms (\textit{right}) on the \chqsumm{} dataset. }
\label{fig:mesh_headings_top_focus}
\end{figure}

\section*{Technical Validation}
\subsection*{Dataset Analysis}
 The \chqsumm{} dataset consists of the $1,507$ original questions and the corresponding summarized questions, question focus, and question types. To benchmark the dataset, we split it into training, validation, and test sets of $1000$, $400$, and $107$ samples, respectively. Table \ref{tab:question-summary-stats} and Table \ref{tab:focus-type-stats} present detailed statistics for the dataset.
The majority of original questions in \chqsumm{} dataset have $200$ words. While most of our human-generated summaries have $15$ words (\textit{cf} Fig \ref{fig:word_distribution}). 
This shows that in a community question-answering forum, users prefer to express their healthcare information needs in overly descriptive, multi-detail posts rather than formulating succinct queries that require more cognitive effort. We also analyzed the distribution of the question focus and question type of the \chqsumm{} dataset. To understand the concept associated with each question focus, we map each focus term to the Medical Subject Headings (MeSH) \cite{mesh_home}.
\paragraph{Finding Medical Subject Headings (MeSH):} 
The Medical Subject Headings (MeSH) is a thesaurus developed by the National Library of Medicine and used for indexing, cataloging, and searching biomedical and health-related information and documents. Since biomedical literature is indexed with MeSH, it is appropriate to use MeSH to search the literature for relevant, reliable answers to consumer healthcare questions.
To this end, we have provided the MeSH headings for the focus of each question in the \chqsumm{} dataset.
We downloaded and pre-processed the Unified Medical Language System\textregistered (UMLS) Metathesaurus from the official UMLS site \cite{bodenreider2004unified,umls_knowledge_sources}.In particular, we utilized the \texttt{MRCONSO.RRF} Metathesaurus file that contains the concepts id, names, codes, etc., from multiple source vocabularies. To map the MeSH headings associated with the question focus, We downloaded and pre-processed the MeSH Descriptor Data (2021) from the official MeSH website \cite{mesh_descriptor_2021}. In the pre-processing step, we build a MeSH dictionary where the MeSH descriptor is stored as  key, and the MeSH tree numbers are stored as the value. 
Given a question focus, we follow the following steps to obtain the MeSH heading:

\begin{figure}[h]
\centering
\begin{subfigure}[b]{0.48\textwidth}
    \centering
    \includegraphics[width=\linewidth,height=6cm]{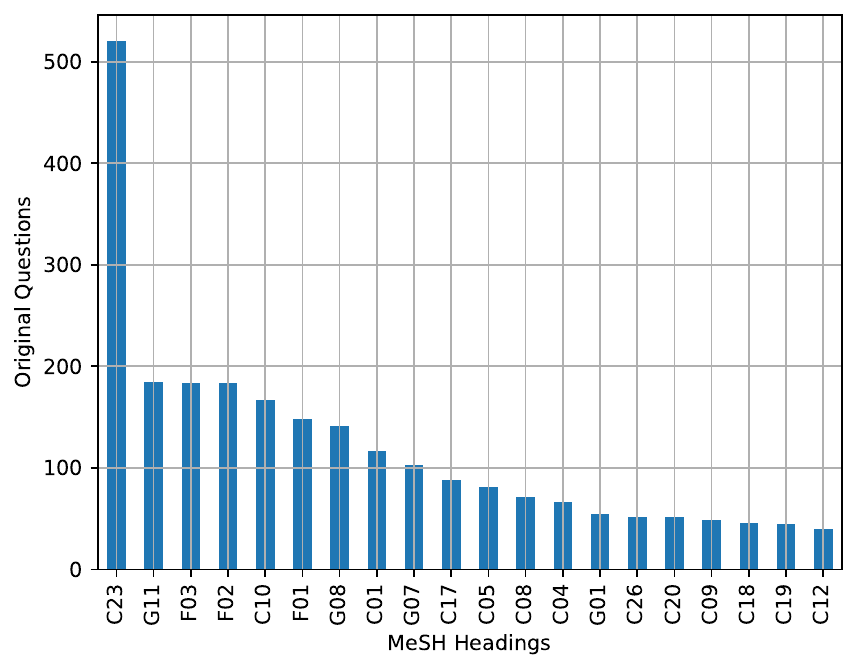}
    \caption{}
    \label{fig:mesh-categories}
\end{subfigure}
\hfill
\begin{subfigure}[b]{0.48\textwidth}
    \centering
    \includegraphics[width=\linewidth,height=6cm]{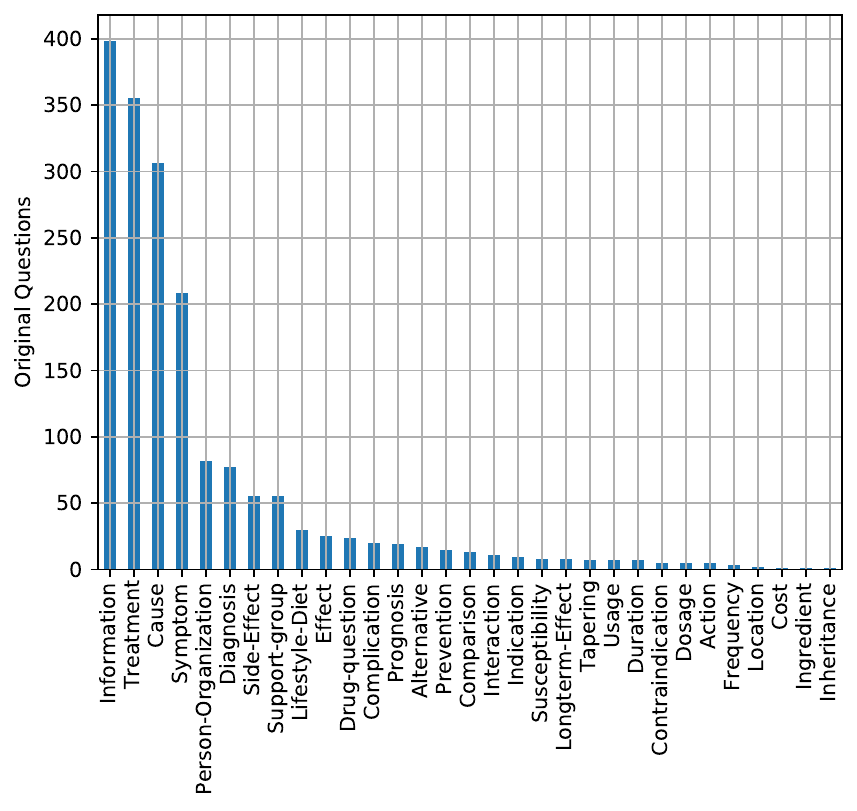}
    \caption{}
\end{subfigure}

\caption{Distribution of the MeSH terms corresponding to the question-focus terms annotated in the original questions of the \textit{CHQ-Summ} dataset (\textit{left}). The \textit{right} side shows the distribution of the different question types annotated in the same dataset.}
\label{fig:focus_and_type_distribution}
\end{figure}

\begin{enumerate}[noitemsep]
    \item We utilized the MetaMap \cite{demner2017metamap,aronson2010overview} tool with the MeSH vocabulary that provides the concept unique identifier (CUI) of the mapping concepts.  
     \item The descriptor unique identifier (DUI) for each mapping CUIs are extracted from  \texttt{MRCONSO.RRF} Metathesaurus. While searching the DUI, we restrict ourselves to only extracting the DUIs for which the language of the term is English (ENG) and the source vocabulary is MeSH (MSH). 
    \item Finally, we used the processed MeSH dictionary to retrieve the associated MeSH Tree numbers and mapped them to their top tree numbers in MeSH Tree.
\end{enumerate}

\begin{figure}[H] 
  \centering
  \begin{subfigure}[b]{0.48\textwidth}
    \centering
    \includegraphics[width=\linewidth,height=5cm]{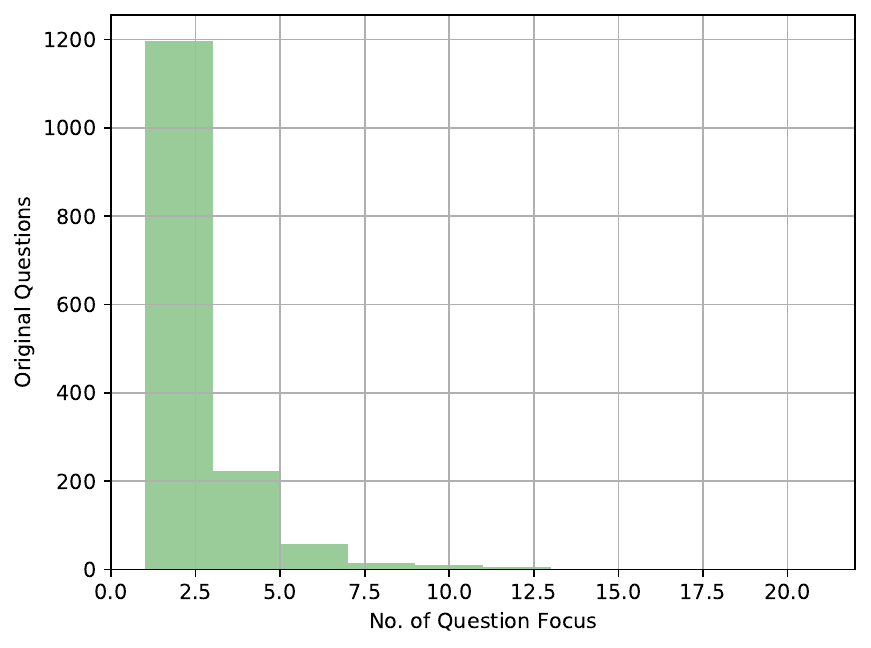}
    \caption{}
  \end{subfigure}\hfill
  \begin{subfigure}[b]{0.48\textwidth}
    \centering
    \includegraphics[width=\linewidth,height=5cm]{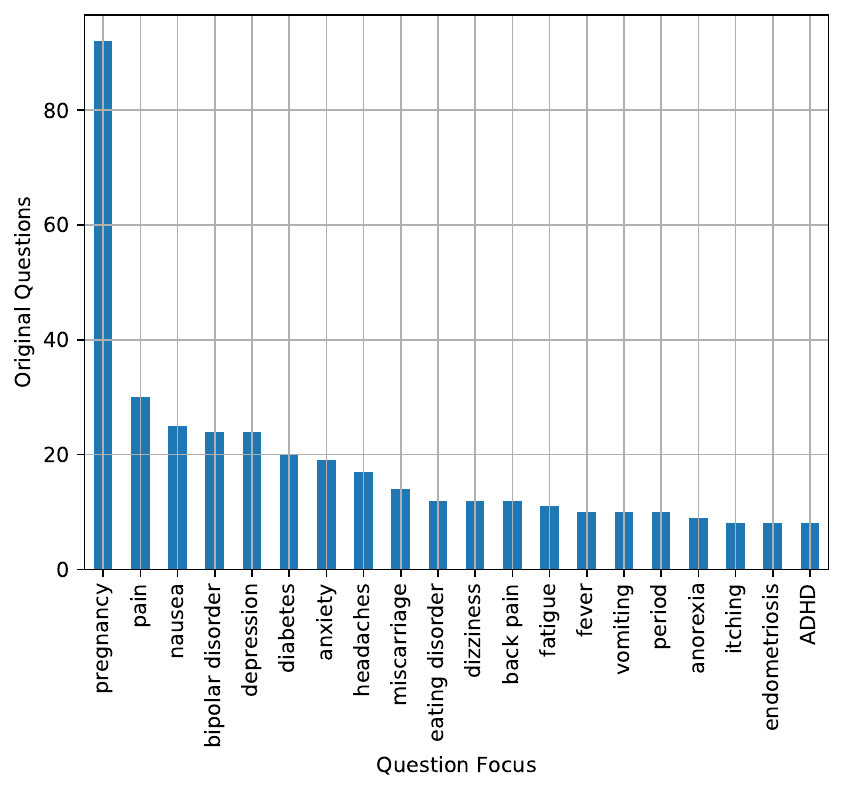}
    \caption{}
  \end{subfigure}
  \caption{Distribution of the number of question-focus terms in the original questions (\textit{left}). Coverage (\textit{right}) of the top-20 question focus on the \textit{CHQ-Summ} dataset.}
  \label{fig:focus_term_and_top_20_focus}
\end{figure}

Using this approach, we identified the top-level MeSH headings associated with the question-focus terms, based on the MeSH Tree numbers defined in the MeSH TreeView \cite{mesh_treeview}.for the question focus present in $1,208$ questions out of total $1,507$ questions in \chqsumm{} dataset. The \chqsumm{} dataset contains a total of $1788$ distinct question focuses and with MetaMap we successfully mapped $1141$ question focuses to MeSH terms. The distribution of top $20$ MeSH terms corresponding to the question-focus terms is provided in Fig-\ref{fig:focus_and_type_distribution} (a). The mapping shows that questions on \textit{``pathological conditions signs and symptoms''} (C23) are most frequent, followed by ``\textit{musculoskeletal and neural physiological phenomena }'' (G11) and ``\textit{mental disorders}'' (F03). The top $20$ MeSH mappings contribute to $96$\% of the total annotated dataset (\textit{cf.} Fig. \ref{fig:mesh_headings_top_focus} (a)). We also analyze the coverage (\textit{cf.} Fig. \ref{fig:mesh_headings_top_focus} (b)) of the most frequent $k$ question focuses and found that top-10, top-20 and top-30 question focuses cover $248$, $321$ and $364$ questions respectively. We noticed that most of the questions in the \chqsumm{} dataset have two focus terms as shown in (cf. Fig. \ref{fig:focus_term_and_top_20_focus} (a)). We have provided the distribution of top-20 question focus term in the (\textit{cf.} Fig. \ref{fig:focus_term_and_top_20_focus} (b)). This shows that the \chqsumm{} contains diverse set of questions covering different question focuses.
We also conducted analysis on the annotated question types. The statistics regarding question type are provided in Fig. \ref{fig:focus_and_type_distribution} (b). The distribution shows that consumers frequently seek general information regarding their healthcare needs. As such,\textit{`information'} question type is the most common question type in \chqsumm{}. Other common question types are `\textit{treatment}', `\textit{cause}', and `\textit{symptoms}'. 

\begin{table}[!ht]
\centering

\scriptsize 
\setlength{\tabcolsep}{0pt}
\renewcommand{\arraystretch}{0.85}

\begin{adjustbox}{max width=\columnwidth} 
\begin{tabular}{@{}p{0.85\columnwidth}@{}} 
\specialrule{.08em}{.08em}{.08em}
\textbf{Question 1:} \textcolor{black}{
\textit{Manic depression perhaps?? I sometimes think of ending it all?? A What are the symptoms, causes \& treatments for this? I have days when I feel I am on top of the world, and other days when I am carrying the weight of the world on my shoulders. I often worry about things incessantly, like money worries, i'm not doing enough with my life etc... I am normally very head-strong, but when I feel depressed, I sometimes wonder if it would be better to end it all? But then I get angry at myself for having such an awful thought, and tell myself off for being like this, only for it to return at some point. Admittedley, thoughts like that are few \& far between. I used to be a really happy chilled out guy, but now I just generally feel more unhappy, than happy, on the whole. Who should I talk to in the UK, in confidence, as I would feel, uncomfortable discussing such matters. Serious answers only please. don't be insensitive.}
} \\
\midrule
\scriptsize
\textbf{Reference:} 
\textcolor{blue}{What are the symptoms, causes, and treatments for manic depression?} \\
\midrule
\scriptsize
\textbf{BART}: \textcolor{blue-violet}{What are the symptoms, causes \& treatments for manic depression?}\\
\scriptsize
\textbf{Llama 3.1 8B}: \textcolor{blue-violet}{What are the symptoms, causes, and treatments for manic-depressive highs, deep lows, constant worry, and rare intrusive thoughts?}\\
\scriptsize
\textbf{Mistral 7B v0.3}: \textcolor{blue-violet}{What explains manic-depression-like highs, depressive drops, money worry, and occasional intrusive thoughts, and how is it treated?}\\
\scriptsize
\textbf{Deepseek 7B}: \textcolor{blue-violet}{What are the symptoms, causes, and treatments for cycles of manic highs, depressive lows, and infrequent suicidal thoughts?}\\
\specialrule{.08em}{.08em}{.08em}
\scriptsize
\textbf{Question 2:} \textcolor{black}{
\textit{Vitamin B12 deficiency question..............? A i have been run down lately.Have had a few blood tests done by my doctor and he has ruled out a few things.i was told by a class mate about it possibly being vitamin b12 deficiency.i am looking it up and it looks like it is quiet possibly it.one thing is that it says a symptom of it is weight loss and loss of appetite.i have had the opposite since its been bothering me.i have major hunger pangs.can anyone tell me if it can happen while having these symptoms as well.everything else seems to make sense except for that.and any information on it in general would be very much appreciated too.i have searched it on search engines and they all seem to have exactly the same wording.just seem to be copied.thank you very much}} \\
\midrule
\scriptsize
\textbf{Reference:} 
\textcolor{blue}{Does Vitamin B12 deficiency cause weight loss and loss of appetite?} \\
\midrule
\scriptsize
\textbf{BART}: \textcolor{blue-violet}{Can vitamin b12 deficiency cause weight loss and loss of appetite?}\\
\scriptsize
\textbf{Llama 3.1 8B}: \textcolor{blue-violet}{Can Vitamin B12 deficiency cause fatigue even without the typical weight loss and loss of appetite?}\\
\scriptsize
\textbf{Mistral 7B v0.3}: \textcolor{blue-violet}{Does Vitamin B12 deficiency always cause weight loss and low appetite, or can symptoms differ?}\\
\scriptsize
\textbf{Deepseek 7B}: \textcolor{blue-violet}{Is it possible to have Vitamin B12 deficiency with hunger instead of loss of appetite?}\\
\specialrule{.08em}{.08em}{.08em}
\scriptsize
\textbf{Question 3:} \textcolor{black}{
\textit{I have chronic stress and anxiety, I am loosing everything in my life, but do not want pills, what can I do? Â I have problems with stress, however it is not just that, but the fact that every time I start with this condition it turns into a huge fear of choking and my mind starts telling me not to eat. The last time it happened I did not eat anything solid for four months and I suffered severe damage in other parts of my body like my stomach and my heart which is worst. This time it is starting again and I am two weeks under this condition. The last time I was using antidepressants and other drugs, but when I tried cutting them the anxiety made me feel worst. This is why I changed my treatment, now I use relaxation exercises with the help on my doctor. The last time it helped me a lot, but this time I think I need more help. I am taking meditation and tai chi courses and I am expecting to take yoga classes as well. The problem is that this is taking away my life, I have doubts on whether I will be cured one day or if it will take so long that everything I have now will be lost. I need help.}} \\
\midrule
\scriptsize
\textbf{Reference:} 
\textcolor{blue}{What are possible non-drug treatments for chronic stress and anxiety?} \\
\midrule
\scriptsize
\textbf{BART}: \textcolor{blue-violet}{What are treatments for chronic stress and anxiety?}\\
\scriptsize
\textbf{Llama 3.1 8B}: \textcolor{blue-violet}{What non-drug treatments can help chronic stress and anxiety that trigger choking fears and long periods of not eating?}\\
\scriptsize
\textbf{Mistral 7B v0.3}: \textcolor{blue-violet}{What non-medication options can manage severe anxiety cycles with choking fear, appetite loss, and major life disruption?}\\
\scriptsize
\textbf{Deepseek 7B}: \textcolor{blue-violet}{What non-drug therapies are effective for chronic stress and anxiety that cause choking fear and difficulty eating?}\\
\specialrule{.08em}{.08em}{.08em}
\scriptsize
\textbf{Question 4:} 
\textit{ Do you think I have it? A I just wanted to know if it sounds like I have anything wrong, like if I may be diabetic or something. First off, I am insanely paranoid about this disease, so it's possible that the hypochondriac in me is having some affect on my outlook on my "symptoms". Well... sometimes I'm kind of thirsty. Not usually, and it's never to the point where I'm like dying or anything. I go to the bathroom 7-9 times in an 18 hour day. (Is that reason for worrying?) I'm sometimes tired, but then again, I get like 5 hours of sleep every night. Isn't being hungry a symptom? Well, I'm never usually that hungry... so that's good, right? I'm not overweight. I have a slight family history of diabetes, but it was only one person... not like my whole family. So do you think I have it?}\\
\midrule
\scriptsize
\textbf{Reference:} \textcolor{blue}{ 
What are the symptoms of diabetes?} \\ 
\midrule
\scriptsize
\textbf{BART}: \textcolor{blue-violet}{What are the symptoms of diabetes and does hypochondriac syndrome have any effect on my health?}\\
\scriptsize
\textbf{Llama 3.1 8B}: \textcolor{blue-violet}{What are the symptoms of diabetes for someone worried about thirst, frequent urination, fatigue, and family history?}\\
\scriptsize
\textbf{Mistral 7B v0.3}: \textcolor{blue-violet}{what are the symptoms of diabetes?}\\
\scriptsize
\textbf{Deepseek 7B}: \textcolor{blue-violet}{What are the common symptoms of diabetes for people concerned about thirst, frequent urination, tiredness, and family history?}\\
\specialrule{.08em}{.08em}{.08em}
\end{tabular}
\end{adjustbox}
\caption{Sample of the summaries from the \chqsumm{} dataset generated by different pre-trained language models \& LLMs.}  
\label{tab:qualitative-analysis}
\end{table}

\subsection*{Benchmarking}

We benchmarked both fine-tuned encoder--decoder models and inference-only instruction-tuned large language models using a unified evaluation protocol to enable fair and consistent comparison. The fine-tuned baselines include BART~\cite{lewis2020bart}, PEGASUS~\cite{zhang2020pegasus}, ProphetNet~\cite{qi2020prophetnet}, and T5-base~\cite{raffel2020exploring}, all trained on the \chqsumm{} dataset using the official train, validation, and test splits. In addition, we evaluated instruction-tuned models Qwen2-7B-Instruct~\cite{qwen22024}, Mistral-7B-Instruct-v0.3~\cite{jiang2023mistral}, Llama-3.1-8B-Instruct, Llama-3.2-3B-Instruct~\cite{touvron2023llama}, Gemma-7B-IT~\cite{gemma2024model}, and DeepSeek-7B-Chat~\cite{deepseek2024}—without further task-specific fine-tuning.

These instruction-tuned models were evaluated using zero-shot, 2-shot, and 5-shot settings across four distinct prompting strategies designed to test different reasoning capabilities:
\begin{enumerate}
    \item \textbf{Standard Generation:} A direct single-turn instruction to rewrite the input as one short medical question (10--15 words).
    \item \textbf{Element-Aware Prompting:} A two-stage strategy where the model first extracts key medical elements (diseases, drugs, tests, symptoms) and then generates a summary strictly incorporating those extracted entities to mitigate hallucination.
    \item \textbf{Hierarchical } Inspired by coarse-to-fine generation methods~\cite{ravaut2023context}, this method splits the input question into text blocks, independently summarizes each block into key medical phrases, and then synthesizes these intermediate outputs into a final concise question.
    \item \textbf{Chain-of-Density (CoD):} Adapted from~\cite{arxiv2309.04269}, this iterative strategy performs five cycles of refinement. In each step, the model identifies "Missing Entities" from the source text and rewrites the summary to include them without increasing the word count, forcing higher information density.
\end{enumerate}

Across both model classes, we employed the same set of automatic evaluation metrics, grouped into reference-based and reference-free categories. Reference-based metrics assess the similarity between the generated summary and the ground truth, including ROUGE-LSum (F1-score)~\cite{lin2004rouge}, METEOR, and BERTScore-F1~\cite{zhang2019bertscore}. Reference-free metrics evaluate the faithfulness and intent of the summary relative to the original source question and include Semantic Coherence and Entailment consistency.

ROUGE scores were computed using the \texttt{rouge-score} library (v0.1.2) with Porter stemming enabled, specifically utilizing the LCS-based sum aggregation (ROUGE-LSum) to account for sentence-level structure. METEOR was calculated using the NLTK implementation with WordNet resources to capture synonymy and morphological variations. BERTScore was calculated using the default English model (RoBERTa-large) via the \texttt{bert\_score} library to capture semantic similarity. For reference-free evaluation, Semantic Coherence was measured using cosine similarity between embeddings from the \texttt{all-MiniLM-L6-v2} SentenceTransformer. Entailment was assessed using a \texttt{roberta-large-mnli} classifier, where the original question served as the premise and the generated summary as the hypothesis; the probability of the "Entailment" class was used as a proxy for medical faithfulness.

For fine-tuned encoder--decoder models, training was conducted for multiple epochs using the AdamW optimizer and validation-based model selection. During inference, both fine-tuned models and instruction-tuned LLMs employed deterministic decoding settings to ensure reproducibility. Specifically, instruction-tuned LLMs used greedy decoding with a repetition penalty of $1.02$ - $1.03$ to prevent loops. All outputs underwent standardized post-processing to remove prompt artifacts (e.g., "FINAL\_QUESTION:" tags) and normalize formatting prior to metric calculation.

\begin{table}[H]
\centering
\scriptsize
\setlength{\tabcolsep}{10pt}
\renewcommand{\arraystretch}{0.95}
\begin{adjustbox}{max width=\textwidth}
\begin{tabular}{l c c c c c c}
\toprule
\multicolumn{6}{c}{\textbf{Sequence-to-Sequence Models}} \\
\cmidrule(lr){1-6}
\textbf{Model} & \textbf{R-L (sum) } & \textbf{METEOR} & \textbf{B.S (F1)} & \textbf{S.Co} & \textbf{Ent.} \\
\midrule
\multirow{1}{*}{BART-Large}
& 0.3329 & 0.2733 & \textcolor{blue}{\underline{\textbf{0.9142}}} & 0.7016 & 0.7467 \\
\multirow{1}{*}{PEGASUS-Large}
& 0.3487 & 0.2923 & 0.9041 & 0.6695 & \textcolor{blue}{\underline{\textbf{0.8294}}} \\
\multirow{1}{*}{ProphetNet-Large-Uncased}
& \textcolor{blue}{\underline{\textbf{0.3596}}} & \textcolor{blue}{\underline{\textbf{0.3034}}} & 0.9102 & 0.6966 & 0.7319 \\
\multirow{1}{*}{T5-Base}
& 0.3448 & 0.2885 & 0.9097 & \textcolor{blue}{\underline{\textbf{0.7307}}} & 0.8198 \\

\bottomrule
\end{tabular}
\end{adjustbox}
\caption{Performance of fine-tuned sequence-to-sequence models on the CHQ-Summ dataset.}
\label{tab:seq2seq_results}
\end{table}

\subsection*{Results and Analysis}

\textbf{Performance of Fine-Tuned Models.}  
Table~\ref{tab:seq2seq_results} reports the results for the fine-tuned sequence-to-sequence models evaluated on CHQ-Summ. Among the reference-based metrics, ProphetNet-Large-Uncased demonstrates the strongest lexical alignment with human summaries, achieving the highest ROUGE-LSum ($0.3596$) and METEOR ($0.3034$) scores. BART-Large follows closely and secures the highest BERTScore-F1 ($0.9142$), indicating strong semantic similarity to the references. regarding reference-free metrics, T5-Base achieves the highest Semantic Coherence ($0.7307$), suggesting superior topical alignment with the source questions. Notably, PEGASUS-Large attains the highest Entailment score ($0.8294$), indicating that it generates the most factually consistent summaries with the lowest likelihood of hallucination among the fine-tuned baselines.

\textbf{Performance of Instruction-Tuned LLMs.}
The results for instruction-tuned LLMs are reported in Table~\ref{tab:chq_llm_fewshot}. Unlike the fine-tuned baselines, the performance of LLMs varied significantly based on the prompting strategy employed.

Regarding reference-based metrics (lexical overlap), the smaller \textbf{Llama-3.2-3B-Instruct} model demonstrated remarkable performance when utilizing the Hierarchical strategy. Specifically, in the 5-shot Hierarchical setting, it achieved the highest overall ROUGE-LSum ($0.3628$) and METEOR ($0.3073$) scores across all models and strategies. This suggests that decomposing the summarization task into intermediate key phrases allows smaller models to effectively mimic the linguistic style of human references. For semantic similarity, \textbf{DeepSeek-7B-Chat} (Hierarchical, 5-shot) achieved the highest BERTScore-F1 ($0.9092$), followed closely by Qwen2-7B-Instruct (Standard, 2-shot) at $0.9091$.

For reference-free metrics (factuality and faithfulness), Qwen2-7B-Instruct emerged as the clear leader. It achieved the highest Entailment score ($0.8623$) using the Element-Aware strategy (0-shot). This result highlights the effectiveness of the Element-Aware approach: by explicitly extracting medical entities before generating the summary, the model significantly reduces hallucinations and ensures the output is logically supported by the source. Furthermore, Qwen2-7B-Instruct also attained the highest Semantic Coherence ($0.7595$) under the Hierarchical strategy (0-shot), demonstrating superior topical retention.

Overall, the results indicate a trade-off between style and safety. The Hierarchical strategy (paired with few-shot examples) drives the best lexical overlap, effectively teaching the model to copy the human reference style. In contrast, the Element-Aware strategy maximizes medical faithfulness (Entailment), making it the preferred choice for safety-critical applications where factual accuracy is paramount.

\begin{table}[H]
\centering
\scriptsize
\setlength{\tabcolsep}{10 pt}
\renewcommand{\arraystretch}{0.5}
\begin{adjustbox}{max width=\textwidth}
\begin{tabular}{l l c c c c c c}
\toprule
\textbf{Strategy} & \textbf{Model} & \textbf{Shots} &
\textbf{R-L (sum)} & \textbf{METEOR} & \textbf{B.S (F1)} &
\textbf{S.Co} & \textbf{Ent.} \\
\midrule
\multirow{21}{*}{Standard}
& \multirow{3}{*}{Mistral-7B-Instruct-v0.3}
& 0 & 0.2769 & 0.2149 & 0.8959 & 0.7004 & 0.7208 \\
&  & 2 & 0.3616 & 0.2516 & 0.9059 & 0.7009 & 0.7446 \\
&  & 5 & 0.3126 & 0.2653 & 0.9070 & 0.6959 & 0.7580 \\

\cmidrule{2-8}
& \multirow{3}{*}{Llama-3.1-8B-Instruct}
& 0 & 0.3090 & 0.2597 & 0.9088 & 0.6571 & 0.6333 \\
&  & 2 & 0.2761 & 0.2371 & 0.9028 & 0.5101 & 0.4590 \\
&  & 5 & 0.3143 & 0.2547 & 0.9088 & 0.6908 & 0.7916 \\

\cmidrule{2-8}
& \multirow{3}{*}{Llama-3.2-3B-Instruct}
& 0 & 0.2802 & 0.2432 & 0.8979 & 0.6650 & 0.6747 \\
&  & 2 & 0.3202 & 0.2681 & 0.9072 & 0.6550 & 0.6138 \\
&  & 5 & 0.2763 & 0.2170 & 0.8960 & 0.6394 & 0.6859 \\

\cmidrule{2-8}
& \multirow{3}{*}{Qwen2-7B-Instruct}
& 0 & 0.2429 & 0.2122 & 0.8894 & 0.7346 & 0.7617 \\
&  & 2 & 0.3413 & 0.2682 & 0.9091 & 0.7119 & 0.8162 \\
&  & 5 & 0.3312 & 0.2570 & 0.9077 & 0.7237 & 0.8148 \\

\cmidrule{2-8}
& \multirow{3}{*}{Gemma-7B-IT}
& 0 & 0.2380 & 0.1844 & 0.8917 & 0.6798 & 0.8329 \\
&  & 2 & 0.2024 & 0.1548 & 0.8809 & 0.6473 & 0.5664 \\
&  & 5 & 0.2101 & 0.1539 & 0.8826 & 0.6638 & 0.6124 \\

\cmidrule{2-8}
& \multirow{3}{*}{DeepSeek-7B-Chat}
& 0 & 0.3136 & 0.2674 & 0.8991 & 0.7444 & 0.8259 \\
&  & 2 & 0.3106 & 0.2539 & 0.9048 & 0.7165 & 0.7952 \\
&  & 5 & 0.3303 & 0.2574 & 0.9070 & 0.7262 & 0.8510 \\

\midrule
\multirow{21}{*}{Chain-of-Density}
&\multirow{3}{*}{Mistral-7B-Instruct-v0.3}
& 0 & 0.2416 & 0.1886 & 0.8790 & 0.7267 & 0.6757 \\
& & 2 & 0.2706 & 0.2159 & 0.8920 & 0.6883 & 0.6736 \\
& & 5 & 0.2786 & 0.2377 & 0.8937 & 0.6791 & 0.6090 \\

\cmidrule{2-8}
&\multirow{3}{*}{Llama-3.1-8B-Instruct}
& 0 & 0.2713 & 0.2238 & 0.8943 & 0.6990 & 0.6754 \\
& & 2 & 0.2294 & 0.1900 & 0.8821 & 0.6025 & 0.4666 \\
& & 5 & 0.2362 & 0.1818 & 0.8877 & 0.6729 & 0.6550 \\

\cmidrule{2-8}
&\multirow{3}{*}{Llama-3.2-3B-Instruct}
& 0 & 0.2791 & 0.2406 & 0.9001 & 0.6965 & 0.6541 \\
& & 2 & 0.2667 & 0.2218 & 0.8951 & 0.6280 & 0.4687 \\
& & 5 & 0.2214 & 0.1903 & 0.8868 & 0.6117 & 0.5107 \\

\cmidrule{2-8}
&\multirow{3}{*}{Qwen2-7B-Instruct}
& 0 & 0.2355 & 0.2006 & 0.8883 & 0.7439 & 0.6995 \\
& & 2 & 0.2943 & 0.2259 & 0.8994 & 0.7144 & 0.7297 \\
& & 5 & 0.2794 & 0.2357 & 0.9009 & 0.7164 & 0.6255 \\

\cmidrule{2-8}
&\multirow{3}{*}{Gemma-7B-IT}
& 0 & 0.2453 & 0.2048 & 0.8785 & 0.7025 & 0.7160 \\
& & 2 & 0.2279 & 0.2106 & 0.8811 & 0.6904 & 0.6844 \\
& & 5 & 0.2392 & 0.1783 & 0.8830 & 0.6689 & 0.6822 \\

\cmidrule{2-8}
&\multirow{3}{*}{DeepSeek-7B-Chat}
& 0 & 0.2862 & 0.2585 & 0.8896 & 0.7456 & 0.8162 \\
& & 2 & 0.2537 & 0.2242 & 0.8962 & 0.7124 & 0.7045 \\
& & 5 & 0.2564 & 0.1872 & 0.8962 & 0.7149 & 0.7358 \\

\midrule
\multirow{21}{*}{Hierarchical}
&\multirow{3}{*}{Mistral-7B-Instruct-v0.3}
& 0 & 0.2347 & 0.1928 & 0.8820 & 0.7233 & 0.7559 \\
& & 2 & 0.2804 & 0.2400 & 0.8917 & 0.7314 & 0.7699 \\
& & 5 & 0.2709 & 0.2197 & 0.8888 & 0.7219 & 0.7067 \\

\cmidrule(lr){2-8}
&\multirow{3}{*}{Llama-3.1-8B-Instruct}
& 0 & 0.2695 & 0.2316 & 0.8994 & 0.6908 & 0.7671 \\
& & 2 & 0.3037 & 0.2695 & 0.9037 & 0.6773 & 0.7005 \\
& & 5 & 0.3122 & 0.2903 & 0.9071 & 0.6779 & 0.7683 \\

\cmidrule(lr){2-8}
&\multirow{3}{*}{Llama-3.2-3B-Instruct}
& 0 & 0.2627 & 0.2044 & 0.8923 & 0.6792 & 0.6078 \\
& & 2 & 0.3066 & 0.2692 & 0.8985 & 0.6805 & 0.6851 \\
& & 5 & \textcolor{blue}{\underline{\textbf{0.3628}}} & \textcolor{blue}{\underline{\textbf{0.3073}}} & 0.9049 & 0.6715 & 0.7435 \\

\cmidrule(lr){2-8}
&\multirow{3}{*}{Qwen2-7B-Instruct}
& 0 & 0.3024 & 0.2446 & 0.8911 & \textcolor{blue}{\underline{\textbf{0.7595}}} & 0.7993 \\
& & 2 & 0.3291 & 0.2838 & 0.9004 & 0.7484 & 0.8490 \\
& & 5 & 0.3258 & 0.2886 & 0.9001 & 0.7384 & 0.8369 \\

\cmidrule(lr){2-8}
&\multirow{3}{*}{Gemma-7B-IT}
& 0 & 0.2595 & 0.2188 & 0.8902 & 0.6960 & 0.8339 \\
& & 2 & 0.2472 & 0.1987 & 0.8802 & 0.6978 & 0.6526 \\
& & 5 & 0.2400 & 0.1915 & 0.8814 & 0.6753 & 0.6622 \\

\cmidrule(lr){2-8}
&\multirow{3}{*}{DeepSeek-7B-Chat}
& 0 & 0.2909 & 0.2184 & 0.8926 & 0.7165 & 0.8064 \\
& & 2 & 0.3090 & 0.2308 & 0.9054 & 0.7015 & 0.7249 \\
& & 5 & 0.3348 & 0.2744 & \textcolor{blue}{\underline{\textbf{0.9092}}} & 0.7093 & 0.8036 \\

\midrule

\multirow{21}{*}{Element-Aware}
&\multirow{3}{*}{Mistral-7B-Instruct-v0.3}
& 0 & 0.2364 & 0.1884 & 0.8829 & 0.6938 & 0.6745 \\
& & 2 & 0.2909 & 0.2523 & 0.8953 & 0.7365 & 0.7413 \\
& & 5 & 0.2929 & 0.2428 & 0.8924 & 0.7356 & 0.7729 \\

\cmidrule(lr){2-8}
&\multirow{3}{*}{Llama-3.1-8B-Instruct}
& 0 & 0.2667 & 0.2165 & 0.8985 & 0.6510 & 0.5720 \\
& & 2 & 0.2583 & 0.2082 & 0.8944 & 0.6147 & 0.5211 \\
& & 5 & 0.2711 & 0.2224 & 0.8943 & 0.6302 & 0.5169 \\

\cmidrule(lr){2-8}
&\multirow{3}{*}{Llama-3.2-3B-Instruct}
& 0 & 0.2109 & 0.1693 & 0.8843 & 0.6189 & 0.5424 \\
& & 2 & 0.2081 & 0.1766 & 0.8836 & 0.5703 & 0.5021 \\
& & 5 & 0.2164 & 0.1791 & 0.8851 & 0.5832 & 0.5473 \\

\cmidrule(lr){2-8}
&\multirow{3}{*}{Qwen2-7B-Instruct}
& 0 & 0.2820 & 0.2294 & 0.8914 & 0.7555 & \textcolor{blue}{\underline{\textbf{0.8623}}} \\
& & 2 & 0.3028 & 0.2457 & 0.9017 & 0.7440 & 0.7941 \\
& & 5 & 0.3058 & 0.2536 & 0.9009 & 0.7310 & 0.8056 \\

\cmidrule(lr){2-8}
&\multirow{3}{*}{Gemma-7B-IT}
& 0 & 0.1129 & 0.0735 & 0.8416 & 0.2625 & 0.3501 \\
& & 2 & 0.0422 & 0.0317 & 0.8219 & 0.1115 & 0.1945 \\
& & 5 & 0.0521 & 0.0382 & 0.8207 & 0.0924 & 0.1561 \\

\cmidrule(lr){2-8}
&\multirow{3}{*}{DeepSeek-7B-Chat}
& 0 & 0.2826 & 0.2303 & 0.8942 & 0.7285 & 0.7357 \\
& & 2 & 0.2812 & 0.2257 & 0.9005 & 0.7100 & 0.7647 \\
& & 5 & 0.3006 & 0.2443 & 0.9035 & 0.7248 & 0.7519 \\

\bottomrule
\end{tabular}
\end{adjustbox}
\caption{Performance comparison of instruction-tuned LLMs across standard and multi-stage prompting strategies (Chain-of-Density, Hierarchical, Element-Aware) with varying few-shot settings. Bold indicates the best score within each strategy--metric group.}
\label{tab:chq_llm_fewshot}
\end{table}

\subsection*{Human Evaluation}

We conducted a human evaluation using the Yadav et al. \cite{yadav2023towards} procedure to supplement the automatic evaluation metrics and overcome the shortcomings of surface-level overlap measures such as ROUGE in capturing semantic fidelity. 
Two independent annotators with a background in biomedical informatics evaluated the model-generated summaries using a random subset of $30$ consumer healthcare questions from the \chqsumm{} dataset. 
The generated summaries were assessed on a 5-point Likert scale ($1 = \text{Poor}$, $5 = \text{Excellent}$) across the following human evaluation dimensions:
\begin{itemize}[noitemsep]
    \item \textbf{Factual Consistency.} 
    This criterion evaluates whether the generated summary preserves the core medical facts and intent of the original consumer question without introducing hallucinated or misleading information. 
    Consistent with previous works \cite{maynez2020faithfulness,pagnoni2021understanding,yadav2023chqsumm}, we employed two complementary sub-measures:
    \begin{enumerate}[label=(\arabic*),noitemsep]
        \item \textit{Factual Correctness:} assesses whether all key medical entities and relations (e.g., diseases, symptoms, treatments) from the original question are preserved accurately in the summary.
        \item \textit{Informativeness:} measures the extent to which the summary retains the overall intent and contextual meaning of the question.
    \end{enumerate}

    \item \textbf{Fluency.} 
    This metric examines grammatical well-formedness and readability of the generated summary. 
    We adopt the DUC fluency guidelines \cite{over2007duc}, where a sentence is considered fluent if it is free from capitalization or grammatical errors and reads naturally without structural disfluencies.
\end{itemize}

\begin{table}[H] 
\centering
\scriptsize
\setlength{\tabcolsep}{9pt}\renewcommand{\arraystretch}{1.1}
\begin{tabular}{l | c c c }
\toprule
\textbf{Model} & \textbf{FC} & \textbf{INF} & \textbf{FL}  \\
\midrule
LLaMA 3.2 3B & \textcolor{blue}{\textbf{4.43}} & \textcolor{blue}{\textbf{4.51}} & \textcolor{blue}{\textbf{4.91}} \\
Qwen 2 7B & 3.41 & 3.54 & 4.26 \\
Deepseek 7B   & 3.98 & 4.09 & 4.81 \\
BART   & \textcolor{blue}{4.14} & \textcolor{blue}{4.18} & \textcolor{blue}{4.85} \\
Pegasus   & 3.39 & 3.63 & 4.54 \\
Prophetnet & 3.84 & 3.85 & 4.11 \\
\bottomrule
\end{tabular}

\caption{Human evaluation on a randomly selected set of 30 CHQ-Summ questions.Metrics: Factual Correctness (FC), Informativeness (INF), and Fluency (FL). }
\label{tab:human_eval_chq}
\end{table}

The human evaluation in Table~\ref{tab:human_eval_chq} highlights the performance differences between lightweight instruction-tuned LLMs and established encoder-decoder baselines. LLaMA 3.2 3B stood out as the superior model. It provided summaries that were the most accurate (FC 4.43), maintained the original question's intent the best (INF 4.51), and achieved the highest fluency (FL 4.91). The encoder-decoder model BART also produced high-quality summaries, ranking second in factual correctness (4.14) and informativeness (4.18).  While all models generated readable text, LLaMA 3.2 3B was uniquely able to maximize fluency without compromising factual consistency. Overall, these findings demonstrate that recent advances in LLMs significantly improve their understanding of real-world consumer health questions, enabling performance beyond traditional baselines.

\subsection*{Evaluation on healthcare answer retrieval task}

To evaluate whether the \textit{summarized consumer health questions (CHQs)} can improve \textit{answer retrieval performance}, we conducted experiments on the LiveQA 2017 test set \citep{abacha2017overview}, which contains 104 medical questions sourced from the National Library of Medicine (NLM). The goal was to retrieve the most relevant and complete answer for each question. For this purpose, we used summarized questions generated by all CHQ summarization models evaluated in this work, alongside the original questions and human-generated (reference) summaries for comparison. 
Answer retrieval was performed using the MedQuAD collection \citep{yadav2022towards}, following the LiveQA evaluation framework, which provides human judgment scores—``\textit{Correct and Complete (4)}'', ``\textit{Correct but Incomplete (3)}'', ``\textit{Incorrect but Related (2)}'', and ``\textit{Incorrect (1)}''. We restricted the evaluation to 48 questions for which consistent human judgment scores were available across all three variants: original, model-generated, and human-generated questions.

We applied the official LiveQA metrics to compare the retrieval performance:
\begin{itemize}[noitemsep,topsep=2pt]
    \item \textbf{avgTop1:} Average judgment score for the top-ranked answer (scaled 0--3).
    \item \textbf{avgBest@4:} Average of the best score within the top-4 retrieved answers.
    \item \textbf{nDCG@4:} Normalized discounted cumulative gain at rank 4.
    \item \textbf{succ@k:} Fraction of questions with a score $\geq k$ (for $k=\{3,4\}$).
    \item \textbf{prec@k:} Fraction of retrieved answers with a score $\geq k$.
    \item \textbf{MRR@k:} Mean reciprocal rank of correct answers at cutoffs 3 and 4.
\end{itemize}

\begin{table}[H]
\centering
\scriptsize
\setlength{\tabcolsep}{2pt}
\renewcommand{\arraystretch}{0.7}
\begin{adjustbox}{max width=\textwidth}
\begin{tabular}{l  l  c c c c c c c c c}
\toprule
Model & Variant & avgTop1 & avgBest@4 & nDCG@4 & succ@3(3+) & succ@4(4+) & prec@3(3+) & prec@4(3+) & MRR@4(3+) & MRR@4(4+)\\
\midrule
\multirow{3}{*}{Gemma-7B}& Original Question  & 0.960 & 1.580 & 0.678 & 0.680 & 0.080 & 0.345 & 0.020 & 0.495 & 0.070 \\
& Reference Summary  & 0.980 & 1.440 & 0.653 & 0.580 & 0.080 & 0.320 & 0.030 & 0.445 & 0.070 \\
& Generated Summary  & 0.980 & 1.400 & 0.659 & 0.600 & 0.040 & 0.290 & 0.010 & 0.490 & 0.025 \\
\midrule
\multirow{3}{*}{Mistral-7B-v0.3} & Original Question  & 0.960 & 1.580 & 0.678 & 0.680 & 0.080 & 0.345 & 0.020 & 0.495 & 0.070 \\
 & Reference Summary  & 0.980 & 1.440 & 0.653 & 0.580 & 0.080 & 0.320 & 0.030 & 0.445 & 0.070 \\
 & Generated Summary  & 1.100 & 1.480 & 0.708 & 0.620 & 0.060 & 0.405 & 0.030 & 0.528 & 0.035 \\
 \midrule
\multirow{3}{*}{Llama-3.1-8B} & Original Question  & 0.960 & 1.580 & 0.678 & 0.680 & 0.080 & 0.345 & 0.020 & 0.495 & 0.070 \\
& Reference Summary  & 0.980 & 1.440 & 0.653 & 0.580 & 0.080 & 0.320 & 0.030 & 0.445 & 0.070 \\
& Generated Summary  & 1.080 & 1.520 & 0.704 & 0.620 & 0.100 & \textcolor{blue}{\underline{\textbf{0.425}}} & \textcolor{blue}{\underline{\textbf{0.055}}} & 0.515 & 0.077 \\
\midrule
\multirow{3}{*}{Llama-3.2-3B}    & Original Question  & 0.960 & 1.580 & 0.678 & 0.680 & 0.080 & 0.345 & 0.020 & 0.495 & 0.070 \\
& Reference Summary  & 0.980 & 1.440 & 0.653 & 0.580 & 0.080 & 0.320 & 0.030 & 0.445 & 0.070 \\
& Generated Summary  & 0.940 & \textcolor{blue}{\underline{\textbf{1.640}}} & 0.648 & \textcolor{blue}{\underline{\textbf{0.720}}} & \textcolor{blue}{\underline{\textbf{0.120}}} & 0.400 & 0.035 & 0.538 & 0.077 \\
\midrule
\multirow{3}{*}{Pegasus-Large}   & Original Question  & 0.960 & 1.580 & 0.678 & 0.680 & 0.080 & 0.345 & 0.020 & 0.495 & 0.070 \\
& Reference Summary  & 0.980 & 1.440 & 0.653 & 0.580 & 0.080 & 0.320 & 0.030 & 0.445 & 0.070 \\
& Generated Summary  & 0.720 & 1.260 & 0.540 & 0.520 & 0.060 & 0.270 & 0.015 & 0.382 & 0.060 \\
\midrule
\multirow{3}{*}{T5-Base}         & Original Question  & 0.960 & 1.580 & 0.678 & 0.680 & 0.080 & 0.345 & 0.020 & 0.495 & 0.070 \\
& Reference Summary  & 0.980 & 1.440 & 0.653 & 0.580 & 0.080 & 0.320 & 0.030 & 0.445 & 0.070 \\
  & Generated Summary  & 0.860 & 1.460 & 0.666 & 0.600 & 0.040 & 0.335 & 0.010 & 0.460 & 0.040 \\
\midrule
\multirow{3}{*}{Qwen2-7B}        & Original Question  & 0.960 & 1.580 & 0.678 & 0.680 & 0.080 & 0.345 & 0.020 & 0.495 & 0.070 \\
 & Reference Summary  & 0.980 & 1.440 & 0.653 & 0.580 & 0.080 & 0.320 & 0.030 & 0.445 & 0.070 \\
 & Generated Summary  & 1.120 & \textcolor{blue}{\underline{\textbf{1.640}}} & 0.717 & \textcolor{blue}{\underline{\textbf{0.720}}} & 0.080 & \textcolor{blue}{\underline{\textbf{0.425}}} & 0.020 & 0.607 & 0.023 \\
\midrule
\multirow{3}{*}{DeepSeek-7B}     & Original Question  & 0.960 & 1.580 & 0.678 & 0.680 & 0.080 & 0.345 & 0.020 & 0.495 & 0.070 \\
   & Reference Summary  & 0.980 & 1.440 & 0.653 & 0.580 & 0.080 & 0.320 & 0.030 & 0.445 & 0.070 \\
    & Generated Summary  & \textcolor{blue}{\underline{\textbf{1.420}}} & \textcolor{blue}{\underline{\textbf{1.640}}} & \textcolor{blue}{\underline{\textbf{0.804}}} & 0.660 & 0.100 & 0.400 & 0.030 & \textcolor{blue}{\underline{\textbf{0.613}}} & \textcolor{blue}{\underline{\textbf{0.100}}} \\
\midrule
\multirow{3}{*}{BART}     & Original Question  & 0.960 & 1.580 & 0.678 & 0.680 & 0.080 & 0.345 & 0.020 & 0.495 & 0.070 \\
   & Reference Summary  & 0.980 & 1.440 & 0.653 & 0.580 & 0.080 & 0.320 & 0.030 & 0.445 & 0.070 \\
    & Generated Summary  & 0.960 & 1.560 & 0.676 & 0.620 & 0.100 & 0.355 & 0.025 & 0.483 & 0.072 \\
    \midrule
\multirow{3}{*}{ProphetNet-Large}     & Original Question  & 0.960 & 1.580 & 0.678 & 0.680 & 0.080 & 0.345 & 0.020 & 0.495 & 0.070 \\
   & Reference Summary  & 0.980 & 1.440 & 0.653 & 0.580 & 0.080 & 0.320 & 0.030 & 0.445 & 0.070 \\
    & Generated Summary  & 0.920 & 1.540 & 0.636 & 0.540 & 0.097 & 0.361 & 0.028 & 0.475 & 0.069 \\
\bottomrule
\end{tabular}
\end{adjustbox}
\caption{LiveQA-based retrieval performance using top-$k$ and ranking metrics. Best scores per metric are highlighted.}
\label{tab:liveqa_transposed}
\end{table}
\noindent

Results from Table~\ref{tab:liveqa_transposed} is to assess whether summarization improves downstream answer retrieval, we evaluated all systems on the LiveQA-based retrieval task using BM25 ranking and an LLM-as-judge scoring framework, where Mistral-7B-Instruct v0.3 was used to assign 1--4 relevance scores for each retrieved answer. The summarized variants, both human-written and model-generated, consistently outperformed original questions across multiple retrieval metrics, confirming the utility of abstraction. Among individual models, DeepSeek-7B achieved the highest top-1 average score (1.420) and the best nDCG@4 (0.804), while both Qwen2-7B and DeepSeek-7B yielded the strongest average best@4 score (1.640). For success-based metrics, Llama-3.2-3B achieved the highest succ@4 (0.120), and Llama-3.1-8B together with Qwen2-7B obtained the highest prec@3+ (0.425). DeepSeek-7B also led in MRR@4 (0.100), reflecting its ability to rank highly relevant answers within the top few retrieved documents. Collectively, these results demonstrate that question summarization whether reference or generated substantially enhances answer retrieval performance, yielding more relevant, higher-ranked medical answers compared to using original consumer health questions.

The results reported in Table~\ref{tab:liveqa_human} evaluate whether summarization improves downstream answer retrieval when relevance is judged by human annotators. We conducted this analysis on a randomly selected set of 30 consumer health questions using a LiveQA-style retrieval setup with BM25 ranking. Human judges assigned relevance scores (from 1 to 4) to the retrieved answers. Under this setting, both human-written summaries and model-generated summaries consistently outperformed the original questions across multiple retrieval metrics.
Among individual models, DeepSeek-7B achieved the highest top-1 average score (1.067) and the strongest MRR@4(3+) (0.547), while Mistral-7B-v0.3 produced the best average best@4 value (1.500) together with an improved nDCG@4 (0.654). For success-based metrics, DeepSeek-7B and Mistral-7B-v0.3 both achieved the highest succ@3 (0.667) and succ@4 (0.100). Qwen2-7B demonstrated competitive precision with the strongest prec@3+ value (0.350) among generated-summary systems. Collectively, these human-scored results show that question summarization whether reference or generated enhances answer retrieval performance and yields more relevant and higher-ranked medical answers relative to using the original consumer questions.

\begin{table}[H]
\centering
\scriptsize
\setlength{\tabcolsep}{2pt}
\renewcommand{\arraystretch}{0.7}
\begin{adjustbox}{max width=\textwidth}
\begin{tabular}{l  l  c c c c c c c c c}
\toprule
Model & Variant & avgTop1 & avgBest@4 & nDCG@4 & succ@3(3+) & succ@4(4+) & prec@3(3+) & prec@4(4+) & MRR@4(3+) & MRR@4(4+)\\
\midrule
\multirow{3}{*}{Gemma-7B} 
& Original Question  & 0.733 & 1.267 & 0.492 & 0.600 & 0.067 & 0.275 & 0.017 & 0.450 & 0.067 \\
& Reference Summary  & 0.933 & 1.267 & 0.525 & 0.533 & 0.100 & 0.317 & 0.042 & 0.447 & 0.100 \\
& Generated Summary  & 0.967 & 1.333 & 0.626 & 0.600 & 0.033 & 0.292 & 0.008 & 0.500 & 0.011 \\
\midrule

\multirow{3}{*}{Mistral-7B-v0.3} 
& Original Question  & 0.733 & 1.267 & 0.492 & 0.600 & 0.067 & 0.275 & 0.017 & 0.450 & 0.067 \\
& Reference Summary  & 0.933 & 1.267 & 0.525 & 0.533 & 0.100 & 0.317 & 0.042 & 0.447 & 0.100 \\
& Generated Summary  & 1.033 & \textcolor{blue}{\underline{\textbf{1.500}}} & \textcolor{blue}{\underline{\textbf{0.654}}} & \textcolor{blue}{\underline{\textbf{0.667}}} & \textcolor{blue}{\underline{\textbf{0.100}}} & 0.408 & 0.033 & 0.531 & 0.067 \\
\midrule

\multirow{3}{*}{Llama-3.1-8B} 
& Original Question  & 0.733 & 1.267 & 0.492 & 0.600 & 0.067 & 0.275 & 0.017 & 0.450 & 0.067 \\
& Reference Summary  & 0.933 & 1.267 & 0.525 & 0.533 & 0.100 & 0.317 & 0.042 & 0.447 & 0.100 \\
& Generated Summary  & 1.000 & 1.400 & 0.621 & 0.633 & 0.067 & \textcolor{blue}{\underline{\textbf{0.350}}} & 0.025 & 0.515 & 0.072 \\
\midrule

\multirow{3}{*}{Llama-3.2-3B} 
& Original Question  & 0.733 & 1.267 & 0.492 & 0.600 & 0.067 & 0.275 & 0.017 & 0.450 & 0.067 \\
& Reference Summary  & 0.933 & 1.267 & 0.525 & 0.533 & 0.100 & 0.317 & 0.042 & 0.447 & 0.100 \\
& Generated Summary  & 0.967 & 1.367 & 0.606 & 0.600 & 0.067 & 0.317 & 0.017 & 0.493 & 0.061 \\
\midrule

\multirow{3}{*}{Pegasus-Large} 
& Original Question  & 0.733 & 1.267 & 0.492 & 0.600 & 0.067 & 0.275 & 0.017 & 0.450 & 0.067 \\
& Reference Summary  & 0.933 & 1.267 & 0.525 & 0.533 & 0.100 & 0.317 & 0.042 & 0.447 & 0.100 \\
& Generated Summary  & 0.833 & 1.233 & 0.506 & 0.567 & 0.067 & 0.283 & 0.017 & 0.443 & 0.056 \\
\midrule

\multirow{3}{*}{T5-Base} 
& Original Question  & 0.733 & 1.267 & 0.492 & 0.600 & 0.067 & 0.275 & 0.017 & 0.450 & 0.067 \\
& Reference Summary  & 0.933 & 1.267 & 0.525 & 0.533 & 0.100 & 0.317 & 0.042 & 0.447 & 0.100 \\
& Generated Summary  & 0.900 & 1.333 & 0.589 & 0.567 & 0.033 & 0.308 & 0.008 & 0.449 & 0.028 \\
\midrule

\multirow{3}{*}{Qwen2-7B} 
& Original Question  & 0.733 & 1.267 & 0.492 & 0.600 & 0.067 & 0.275 & 0.017 & 0.450 & 0.067 \\
& Reference Summary  & 0.933 & 1.267 & 0.525 & 0.533 & 0.100 & 0.317 & 0.042 & 0.447 & 0.100 \\
& Generated Summary  & 1.033 & 1.400 & 0.620 & 0.600 & 0.067 & \textcolor{blue}{\underline{\textbf{0.350}}} & 0.017 & 0.538 & 0.039 \\
\midrule

\multirow{3}{*}{DeepSeek-7B} 
& Original Question  & 0.733 & 1.267 & 0.492 & 0.600 & 0.067 & 0.275 & 0.017 & 0.450 & 0.067 \\
& Reference Summary  & 0.933 & 1.267 & 0.525 & 0.533 & 0.100 & 0.317 & 0.042 & 0.447 & 0.100 \\
& Generated Summary  & \textcolor{blue}{\underline{\textbf{1.067}}} & \textcolor{blue}{\underline{\textbf{1.500}}} & 0.638 & \textcolor{blue}{\underline{\textbf{0.667}}} & \textcolor{blue}{\underline{\textbf{0.100}}} & 0.383 & 0.025 & \textcolor{blue}{\underline{\textbf{0.547}}} & 0.078 \\
\midrule

\multirow{3}{*}{BART-Large} 
& Original Question  & 0.733 & 1.267 & 0.492 & 0.600 & 0.067 & 0.275 & 0.017 & 0.450 & 0.067 \\
& Reference Summary  & 0.933 & 1.267 & 0.525 & 0.533 & 0.100 & 0.317 & 0.042 & 0.447 & 0.100 \\
& Generated Summary  & 0.900 & 1.333 & 0.573 & 0.567 & 0.067 & 0.292 & 0.017 & 0.445 & 0.056 \\
\midrule

\multirow{3}{*}{ProphetNet-Large} 
& Original Question  & 0.733 & 1.267 & 0.492 & 0.600 & 0.067 & 0.275 & 0.017 & 0.450 & 0.067 \\
& Reference Summary  & 0.933 & 1.267 & 0.525 & 0.533 & 0.100 & 0.317 & 0.042 & 0.447 & 0.100 \\
& Generated Summary  & 0.900 & 1.333 & 0.580 & 0.567 & 0.067 & 0.292 & 0.017 & 0.446 & 0.056 \\
\bottomrule
\end{tabular}
\end{adjustbox}

\caption{Retrieval performance of different models on the healthcare answer-retrieval task using LiveQA judgments. Metrics are computed over all questions with consistent human scores and evaluated using top-$k$ and ranking-based measures.}
\label{tab:liveqa_human}
\end{table}
\noindent

\section*{Data and Code Availability}

We provide detailed instructions in the README file of the Open Science Framework repository describing how to process the \chqsumm{} dataset \cite{osf_chqsumm}. The source code used for all benchmarking experiments is publicly available in a GitHub repository \cite{chqsumm_script}. Due to Yahoo copyright restrictions, we cannot directly distribute the original Yahoo questions. However, the questions are publicly available through the Yahoo! Answers L6 corpus and can be accessed after signing the required Yahoo research agreements \cite{yahoo_l6}.

\section*{Acknowledgements} 
This work was supported by the intramural research program at the U.S. National Library of Medicine, National Institutes of Health, and utilized the computational resources of the NIH High-Performance Computing Biowulf cluster \cite{nih_biowulf}. The content is solely the responsibility of the authors and does not necessarily represent the official views of the National Institutes of Health. We would like to also thank Asma Ben Abacha, Khalil Mrini and Susan Koenig for their annotations and contributions to the pilot stages of this work.



\bibliography{sample}
\end{document}